\documentclass{article}

\usepackage{arxiv}

\usepackage[utf8]{inputenc}
\usepackage[T1]{fontenc}
\usepackage{hyperref}
\hypersetup{
  hidelinks,  % Hide all borders/boxes around links
  colorlinks=true,  % Use colored text instead of boxes (optional, but recommended)
  linkcolor=blue,
  urlcolor=blue,
  pdfauthor={Payam Latifi},
  pdftitle={Sample Document with ORCID}
}

\usepackage{url}
\usepackage{booktabs}
\usepackage{amsfonts}
\usepackage{nicefrac}
\usepackage{microtype}
\usepackage{graphicx}
\usepackage{caption}
\usepackage{longtable} % For multi-page tables
\usepackage{adjustbox}
\graphicspath{{./images/}}
\usepackage{float}
\usepackage{listings}
\usepackage{xcolor}
\usepackage{orcidlink}
\usepackage{orcidlink} % For ORCID icon and link

\usepackage{fancyhdr}
\title{Is 'Hope' a Person or an Idea? A Pilot Benchmark for NER: Comparing Traditional NLP Tools and Large Language Models on Ambiguous Entities}
\author{
  Payam Latifi\,\orcidlink{0009-0006-2369-2785} \\
  Department of Humanities \\
  University of Turin \\
  Turin, Italy \\
  \texttt{Payam.latifi@edu.unito.it}
}

\begin{document}

\maketitle

\begin{abstract}
This pilot study presents a small-scale but carefully annotated benchmark of Named Entity Recognition (NER) performance across six systems: three non-LLM NLP tools (NLTK, spaCy, Stanza) and three general-purpose large language models (LLMs: Gemini-1.5-flash, DeepSeek-V3, Qwen-3-4B). The dataset contains 119 tokens covering five entity types (PERSON, LOCATION, ORGANIZATION, DATE, TIME). We evaluated each system’s output against the manually annotated gold standard dataset using F1-score. The results show that LLMs generally outperform conventional tools in recognizing context-sensitive entities like person names, with Gemini achieving the highest average F1-score. However, traditional systems like Stanza demonstrate greater consistency in structured tags such as LOCATION and DATE. We also observed variability among LLMs, particularly in handling temporal expressions and multi-word organizations. Our findings highlight that while LLMs offer improved contextual understanding, traditional tools remain competitive in specific tasks, informing model selection.
\end{abstract}


\section{Introduction}
Named Entity Recognition (NER) is a fundamental task in Natural Language Processing (NLP), aiming to identify and classify named entities such as persons, locations, organizations, dates, and time expressions within unstructured text. NER plays a crucial role in downstream applications like information extraction, question answering, and semantic search, making it an essential building block for modern language technologies \cite{manning2014stanford}.

In recent years, NER systems have evolved significantly, ranging from rule-based and statistical models to deep learning and transformer-based architectures \cite{devlin2019bert}. Comprehensive reviews of earlier techniques appear in \cite{li2020survey}. Recent surveys highlight LLM potential for few-shot NER \cite{wang2023zero}; the present study employs single-shot prompting for each LLM. Non-LLM NLP tools such as NLTK, spaCy, and Stanza provide efficient, deterministic solutions that are widely used in practical applications due to their speed and reliability. However, the emergence of large language models (LLMs) like Gemini, DeepSeek, and Qwen has introduced new possibilities for more context-aware and flexible entity recognition. This paper addresses: What performance patterns emerge when comparing traditional NLP libraries and LLMs on a small, ambiguity-rich dataset, and how do these inform task-specific model selection?

This paper presents an exploratory, small-scale comparative analysis of non-LLM NER tools and LLMs, focusing on their performance across five key entity types: PERSON, LOCATION, ORGANIZATION, DATE, and TIME. We manually annotated a small but representative dataset and evaluated each tool's ability to align with human annotation using standard evaluation metrics such as accuracy, precision, recall, and F1-score.

Our methodology includes: Manual annotation of a custom dataset, Token-level alignment of model outputs, Quantitative evaluation of six NER systems, Visual and statistical comparison of performance across models and entity types. Given the deliberately small scale of our dataset (119 tokens), the study is framed as an exploratory pilot, prioritizing qualitative insights and reproducibility over large-scale generalization. This work’s novelty lies in its focus on ambiguity (e.g., ‘Hope’ as PERSON) in a small dataset, enabling detailed error analysis unattainable in large corpora. Because the sample contains only 119 tokens (56 entities), numerical differences should be interpreted as directional only; we report no statistical significance tests.

\section{Data}

\subsection{Source and Collection}
The dataset used in this study consists of manually selected and annotated sentences focusing on named entity recognition (NER) tasks. The input text was compiled from various fictional and descriptive narratives, including event descriptions and character interactions. While not sourced from a public corpus or digital archive, the dataset was constructed to reflect real-world NER tagging challenges such as ambiguity, multi-word entities, and overlapping entity types. The dataset was tokenized with NLTK’s Punkt \verb|word_tokenize|, resulting in 119 tokens, which served as the reference tokenization for all systems. All model outputs were manually aligned to this token list to ensure consistent evaluation across tools. The full input text used for evaluation is shown below:

\begin{quote}
\itshape On a chilly Thursday morning, April Blake met Justice Hope at Madison Square, near Washington Tower. At 8:45, a memo from the Center for Civic Leadership reached the office of River Clark, head of Northern Union Trust. Later, Horizon filed a complaint dated October 2018 with the Department of Historical Records. By midday, journalists from The Chronicle and Liberty Press were circling City Hall. Jordan Reed mentioned a meeting on the fourth Friday of March. Meanwhile, Trinity Wells, formerly of Bridgewater School, was seen near Lincoln. Just before dusk, the Mayor’s Office received a tip linked to Ashley Fields and Pine.
\end{quote}

\subsection{Annotation Process}
To ensure a reliable gold standard for comparison, all tokens were manually annotated by the author based on established NER conventions. The annotation followed these principles:
\begin{description}
    \item[PERSON] Names of individuals (e.g., April, Jordan Reed)
    \item[LOCATION] Geographic locations or place names (e.g., Madison Square, Washington Tower, Lincoln, City Hall)
    \item[ORGANIZATION] Organizations, institutions, or companies (e.g., Center for Civic Leadership, Northern Union Trust)
    \item[DATE] Specific calendar dates or named time periods (e.g., October 2018, the fourth Friday of March)
    \item[TIME] Expressions referring to specific times of day (e.g., 8:45, dusk)
\end{description}
Each token was labeled individually, ensuring alignment with tokenization standards used by NLP tools (see Appendix C).

\subsection{Dataset Characteristics}
Although compact, the dataset is engineered to maximize diagnostic value: it packs 119 tokens with exactly the kinds of surface-level traps that historically expose gaps between rule-based, neural and prompt-based taggers. Ambiguous person names such as “Justice” and “Hope” force systems to decide whether a capitalized common noun denotes a role or an individual, a distinction on which lexicon-driven models often flip. Multi-token spans (“Madison Square”, “Northern Union Trust”, “the fourth Friday of March”) test whether taggers correctly open and close brackets across stop-words and prepositional phrases; single-token oversights here cascade into four or five separate errors under token-level evaluation. Finally, context-sensitive temporal expressions (“midday”, “dusk”) require pragmatic, document-level inference rather than gazetteer lookup, penalizing models that rely solely on explicit date strings such as “October 2018”. By concentrating these phenomena in one short passage we can observe boundary-decision and sense-disambiguation strategies side-by-side without the noise of a larger corpus.

\subsection{Entity Distribution}

\begin{table}
\caption{Distribution of Entity Types in the Annotated Dataset (n = 119 tokens)}
\centering
\begin{tabular}{lll}
\toprule
\textbf{Entity Type} & \textbf{Count} & \textbf{Percentage} \\
\midrule
DATE & 7 & 5.88\% \\
LOCATION & 8 & 6.72\% \\
ORGANIZATION & 22 & 18.49\% \\
PERSON & 13 & 10.92\% \\
TIME & 6 & 5.04\% \\
O (non-entities) & 63 & 52.94\% \\
\midrule
Total Tokens & 119 & 100\% \\
\bottomrule
\end{tabular}
\label{tab:entity_dist}
\end{table}

The observed distribution of entities reflects typical patterns found in real-world based corpora, where organizational entities are abundant while temporal expressions are relatively scarce. Across the 119 tokens, non-entity tokens (O) constitute 63 instances (52.9\%), while the remaining 56 tokens are classified as entities. Among these, ORGANIZATION is the most frequent category, accounting for 22 tokens (18.5\%), largely due to multi-word institutional names such as ``Center for Civic Leadership'' and ``Northern Union Trust''. PERSON entities appear 13 times (10.9\%), including both unambiguous names like ``April Blake'' and ambiguous cases such as ``Justice Hope''. LOCATION comprises eight tokens (6.7\%), covering multi-word place names like ``Madison Square'' and ``Washington Tower'', as well as single toponyms such as ``Lincoln''. DATE contributes seven tokens (5.9\%), spanning both explicit references (e.g., ``October 2018'') and phrasal expressions (e.g., ``the fourth Friday of March''). Finally, TIME is the least represented category, with six tokens (5.0\%), including precise readings like ``8:45'' and contextual expressions such as ``midday'' and ``dusk''.

This skewed distribution deliberately challenges NER systems, as it involves complex boundary decisions and extended multi-token spans.

\section{Experiments}

\subsection{Tools and Systems Evaluated}
This study compares six Named Entity Recognition (NER) systems, divided into two main categories:

\subsubsection{Off-the-shelf NLP Libraries}
These are rule-based or machine learning-based systems trained on standard corpora and widely used for NER tasks due to their speed and reliability.

\begin{itemize}
\item \textbf{NLTK}: A foundational NLP library using pre-trained models for tagging \cite{bird2009nltk}.
    \item \textbf{spaCy}: A fast and efficient industrial-strength NLP system with pre-trained pipelines for English \cite{honnibal2020spacy}.
    \item \textbf{Stanza}: A deep learning-based toolkit developed by Stanford, offering high accuracy in multilingual NLP tasks \cite{peng2019stanza}.
\end{itemize}

\subsubsection{General-purpose Large Language Models}
These models use transformer architectures and are capable of context-aware entity recognition without explicit training on domain-specific data.
\begin{itemize}
\item \textbf{Gemini-1.5-Flash}: Google's advanced language model with strong reasoning and NLP capabilities \cite{gemini2024}, which was queried through Google’s official Python SDK on 14 May 2025 with temperature 0.2 and \texttt{max\_tokens} left at the model’s default; the exact prompt is shown in Appendix A.
    \item \textbf{DeepSeek-V3}: A commercial LLM known for its performance in multilingual settings \cite{deepseek2024}. It was reached through OpenRouter on 14 May 2025 at temperature 0.1 and 1000 \texttt{max\_tokens}; the JSON-style prompt is reproduced in Appendix A.
    \item \textbf{Qwen-3-4B}: Alibaba Cloud's large-scale language model designed for a wide range of NLP tasks \cite{yang2025qwen}, and was called via the same OpenRouter endpoint on 14 May 2025 with identical hyper-parameters and prompt as DeepSeek-V3; details are in Appendix A.
\end{itemize}

\subsection{Prompting}
Single-shot prompting was chosen deliberately to mirror the zero-friction setup practitioners use when they quickly swap an LLM into an existing NER pipeline. By giving each model exactly one annotated example wrapped in a short, imperative prompt, we retained the ``out-of-the-box'' behavior that users would encounter without costly prompt engineering or few-shot demonstrations. This design keeps the experiment inexpensive (no extra tokens for long contexts), reproducible (a single fixed prompt per model), and fair across systems: every LLM receives the same concise instruction, eliminating the variance that longer chains or dynamic examples would introduce.

Temperatures were tuned per model to respect each provider’s default personality while keeping outputs as deterministic as possible, however, any temperature $> 0$ retains sampling randomness, so observed differences could partly reflect sampling variation rather than intrinsic capability. Gemini-1.5-flash was set to 0.2—low enough to suppress creativity yet slightly higher than the minimum to avoid the repetitive token loops that Gemini can fall into at $t = 0$. DeepSeek-V3 and Qwen-3-4B were both run at 0.1, the lowest temperature we could set reliably via the OpenRouter endpoint at the time of writing, further reducing sampling noise for the JSON-generation task.

Finally, every sentence was submitted only once. Because all three models are non-deterministic even at low temperature, multiple rolls would have introduced random within-model variance and made cross-model comparison unfair; a single call per text guarantees that any performance gap reflects genuine model differences rather than sampling luck.

\subsection{Model Specifications}
Gemini-1.5-flash (checkpoint: gemini-1.5-flash-latest) was queried via the google-generativeai Python SDK on 14 May 2025. We used the exact prompt shown in Appendix A: a single-turn instruction that lists the five allowed tags, gives comma-separated examples, and ends with the sentence to be tagged. Temperature was set to 0.2 and max\_tokens was left at the model’s default; the model therefore produced as many tokens as needed for the list format.

DeepSeek-V3 was invoked through the OpenRouter endpoint \url{https://openrouter.ai/api/v1/chat/completions} on 14 May 2025. The exact user prompt (Appendix A) asks for a JSON list of objects with keys "text" and "label" and provides an illustrative example; the system message simply states ``You are an expert Named Entity Recognition system...''. Generation parameters were temperature = 0.1 and max\_tokens = 1000.

Qwen-3-4B (qwen/qwen3-4b:free) was accessed through the same OpenRouter endpoint on 14 May 2025. It received the same system message and user prompt as DeepSeek-V3 (see Appendix A), ensuring a fair comparison. Temperature was kept at 0.1 and max\_tokens = 1000; no repetition or sampling was applied.

\subsection{Data Collection and Token Alignment}
Each tool was run independently on the same input text used in our dataset. Outputs were collected in CSV format and token-level alignment was performed using NLTK as the reference tokenization, ensuring consistency across all systems.

Some tools (especially LLMs like Gemini, DeepSeek, and Qwen) returned multi-token entity predictions, requiring manual distribution of labels across individual tokens to match NLTK’s tokenization.

\subsection{Evaluation Metrics}
The performance of each system was evaluated against the human-annotated gold standard using four widely adopted metrics: accuracy, precision, recall, and F1-score. Among these, the F1-score was selected as the primary evaluation metric due to its effectiveness in handling imbalanced datasets. It was computed for each entity type individually and subsequently macro-averaged to provide a comprehensive comparison of overall system performance. Macro-averaging includes entity types whose F1 was 0, thereby lowering the global mean.

\section{Results and Analysis}

\subsection{Per-Tool Performance by Entity Type}

\begin{figure}
  \centering
      \includegraphics[width=10cm]{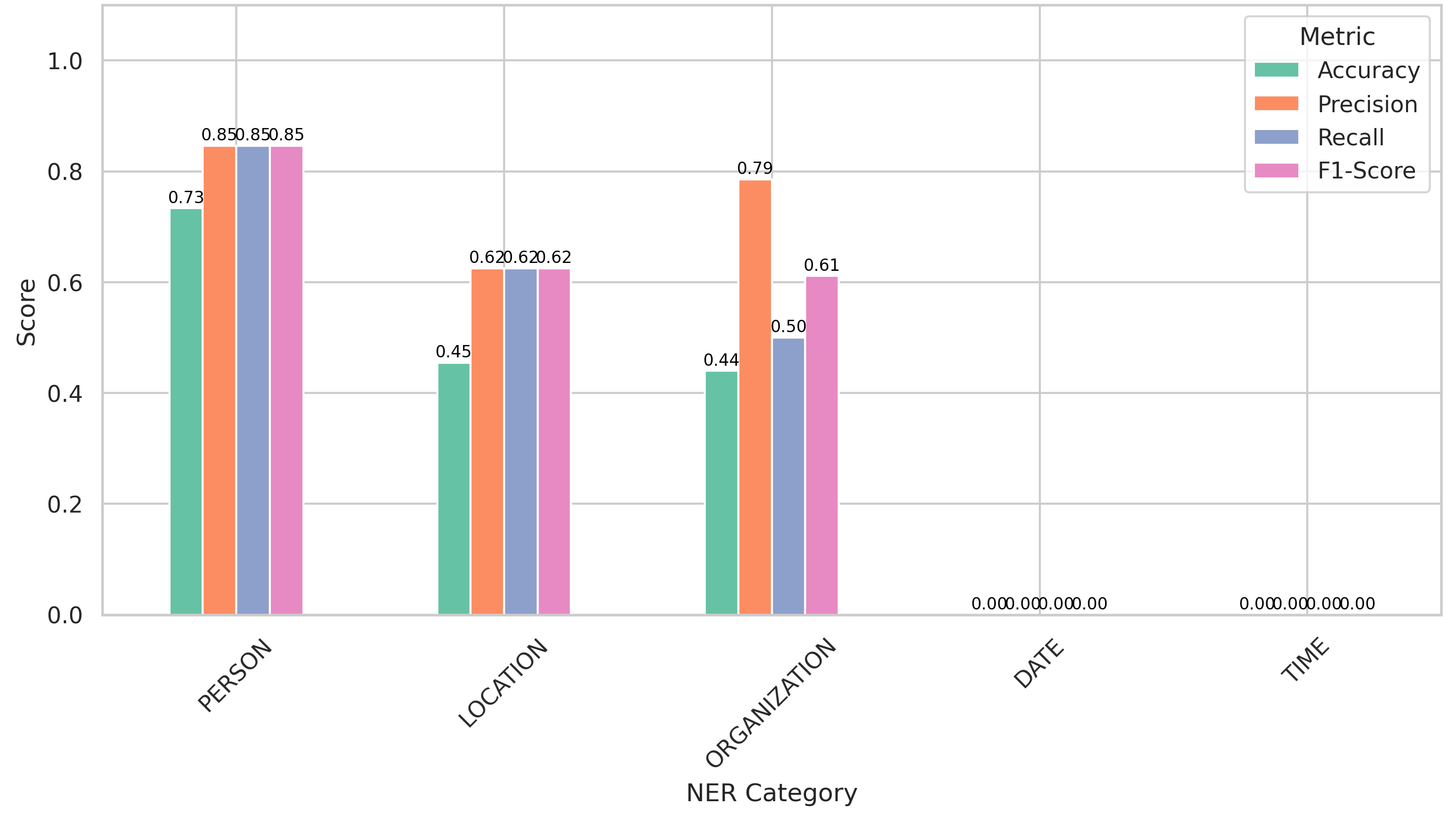} % Replace with your actual filename
  \caption{Per-Entity F1-Score for NLTK Across Five Entity Types}
  \label{fig:nltk}
\end{figure}

As shown in Figure~\ref{fig:nltk}, NLTK demonstrates moderate performance in Named Entity Recognition. It performs best at identifying PERSON entities (F1 = 0.846), showing reasonable accuracy in recognizing individual names like April and Blake. However, it struggles significantly with temporal expressions (both DATE and TIME are assigned an F1-score of 0.0, indicating that no correct predictions were made in these categories). Its performance on LOCATION (F1 = 0.625) and ORGANIZATION (F1 = 0.611) is limited, suggesting that NLTK lacks contextual understanding needed to correctly identify spatial or institutional names consistently.

\begin{figure}
  \centering
    \includegraphics[width=10cm]{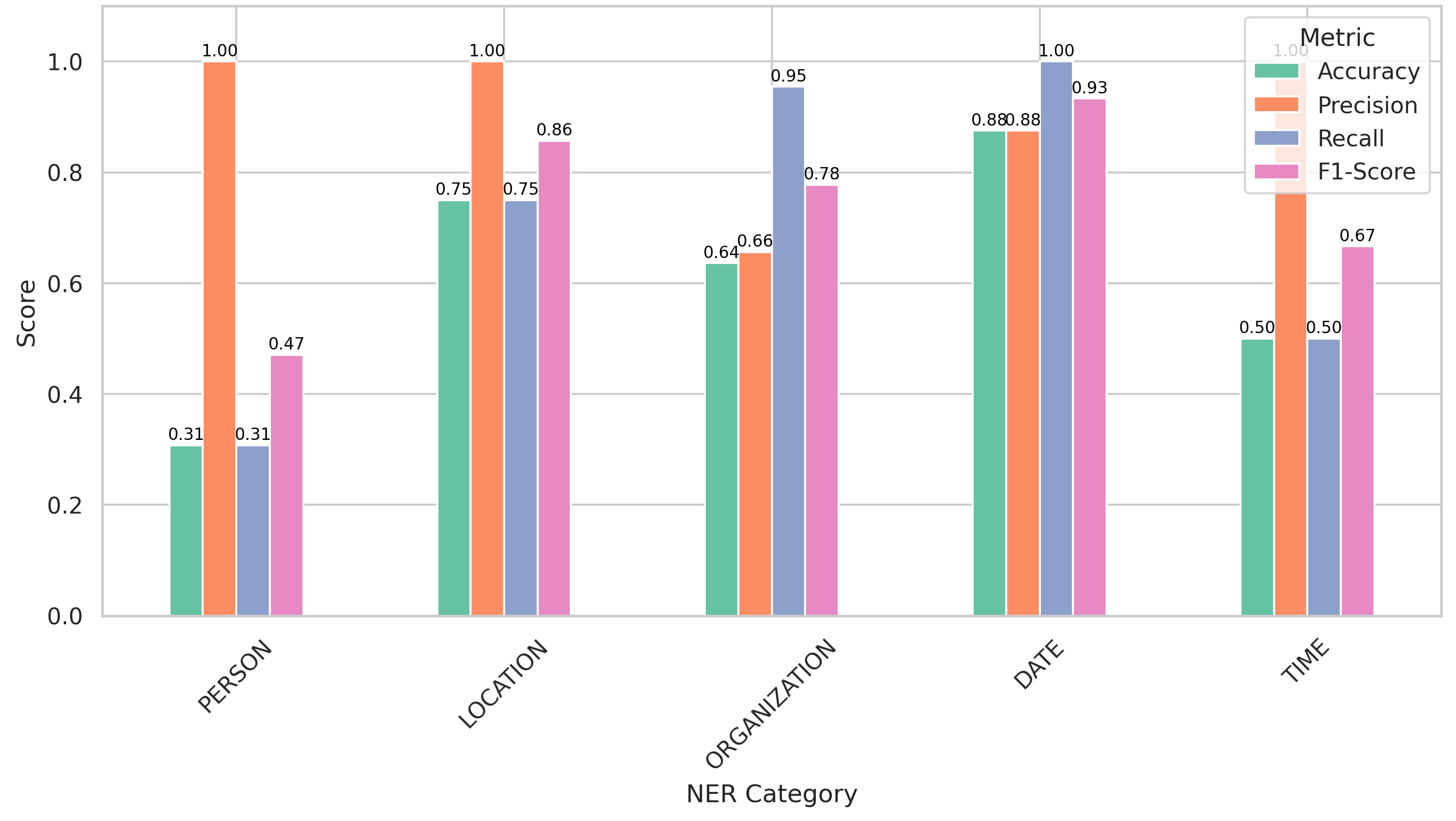} % Replace with your actual filename
  \caption{Per-Entity F1-Score for spaCy Across Five Entity Types}
  \label{fig:spacy}
\end{figure}

spaCy shows a balanced performance across most entity types. It achieves the highest score in DATE recognition (F1 = 0.933), demonstrating strong capability in identifying calendar-based expressions such as October and 2018. It also performs well on LOCATION (F1 = 0.857) and ORGANIZATION (F1 = 0.778). However, spaCy struggles with ambiguous person names like Justice and Hope, where it often labels them as O; its PERSON precision reaches 1.0 (zero false positives) but recall is only 0.307, yielding an F1-score of 0.471 and indicating room for improvement in context-sensitive tagging. It performs reasonably on TIME (F1 = 0.667), but not as strongly as some other tools (Figure~\ref{fig:spacy}).

\begin{figure}
  \centering
    \includegraphics[width=10cm]{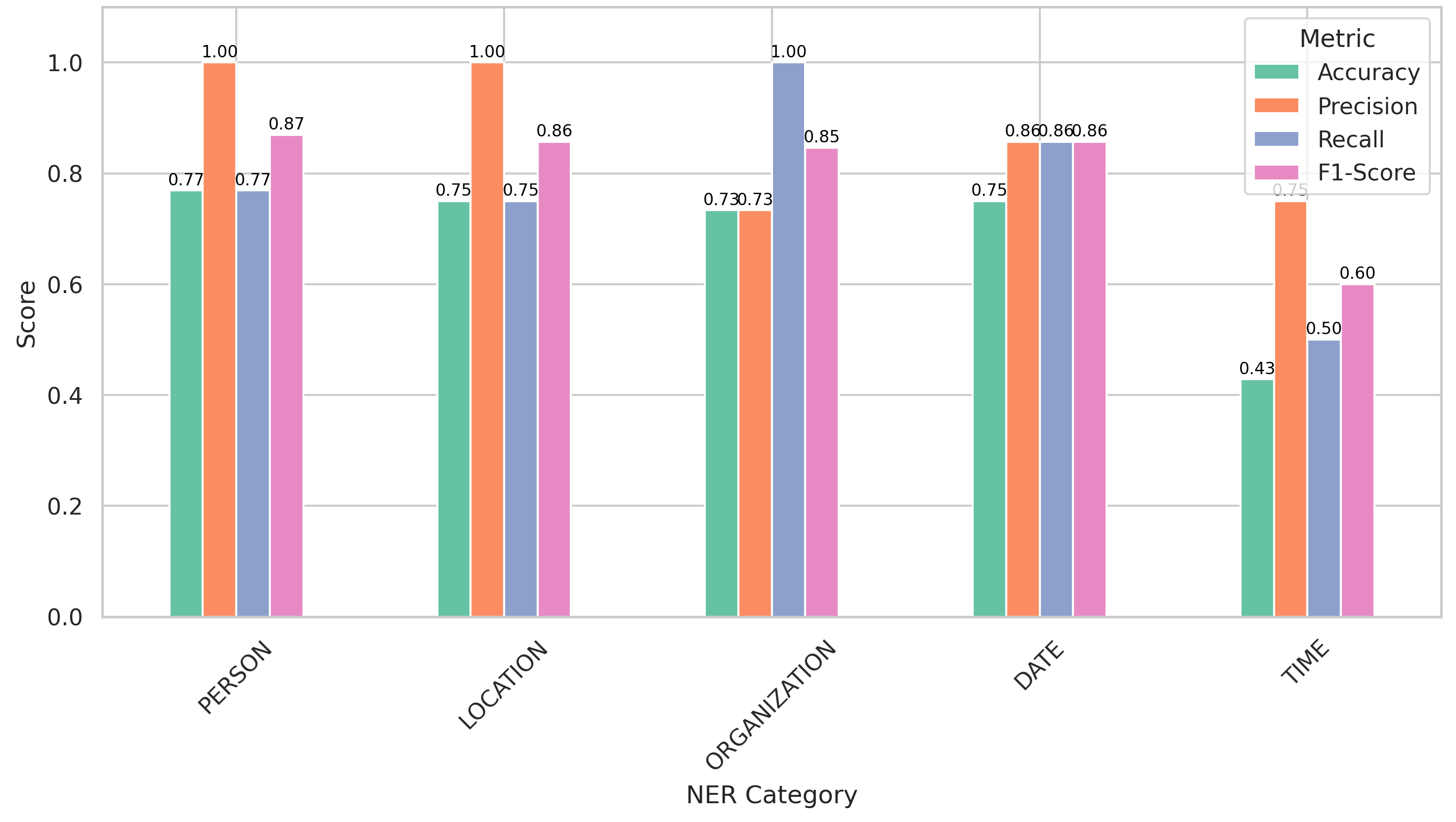} % Replace with your actual filename
  \caption{Per-Entity F1-Score for Stanza Across Five Entity Types}
  \label{fig:stanza}
\end{figure}

Stanza delivers consistently high performance across all categories. It excels at recognizing PERSON (F1 = 0.870), outperforming both NLTK and spaCy. It matches spaCy’s performance on LOCATION (F1 = 0.857) and slightly surpasses it in ORGANIZATION tagging (F1 = 0.846). Stanza also maintains solid results on temporal expressions: DATE (F1 = 0.857) and TIME (F1 = 0.600). These results suggest that Stanza's deep learning-based pipeline provides better contextual awareness than rule-based systems like NLTK (Figure~\ref{fig:stanza}).

% \begin{figure}
%   \centering
%   \fbox{\rule[-.5cm]{10cm}{5cm}}
%   \caption{Per-Entity F1-Score for Gemini-1.5-flash Across Five Entity Types}
%   \label{fig:gemini}
% \end{figure}

\begin{figure}
  \centering
  \includegraphics[width=10cm]{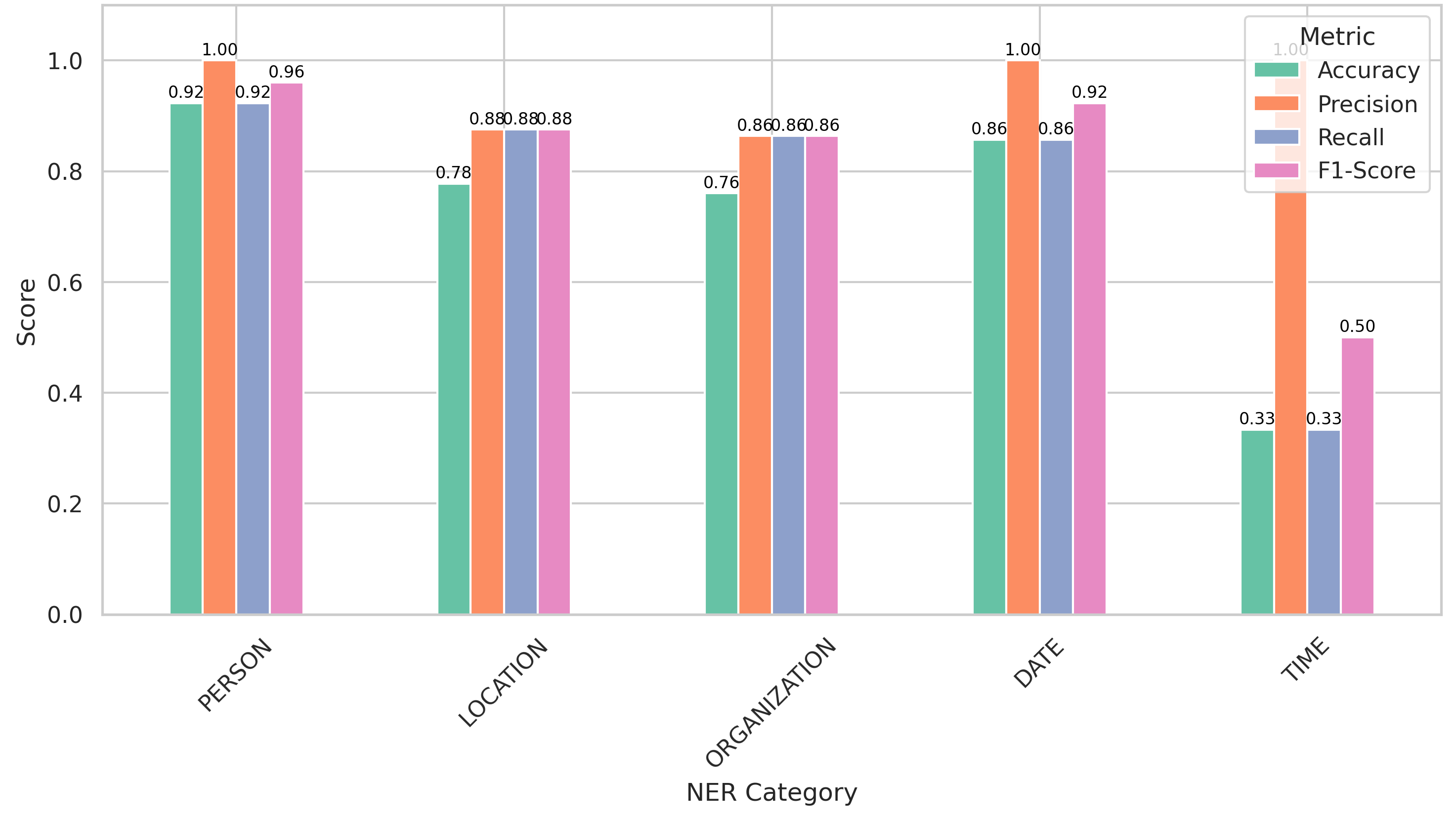} % Replace with your actual filename
  \caption{Per-Entity F1-Score for Gemini-1.5-flash Across Five Entity Types}
  \label{fig:gemini}
\end{figure}

Gemini achieves the highest overall performance among all evaluated tools. It scores perfectly on PERSON recognition (F1 = 0.960), accurately labeling ambiguous names like Justice and Hope as people. It also performs very well on LOCATION (F1 = 0.875) and DATE (F1 = 0.923), making it highly effective at recognizing both spatial and temporal references. However, its performance on TIME (F1 = 0.500) is notably weaker compared to its other results, suggesting limitations in handling time-of-day expressions like dusk or midday. Overall, Gemini’s ability to understand context makes it particularly strong in NER tasks involving real-world ambiguity (Figure~\ref{fig:gemini}).

\begin{figure}
  \centering
      \includegraphics[width=10cm]{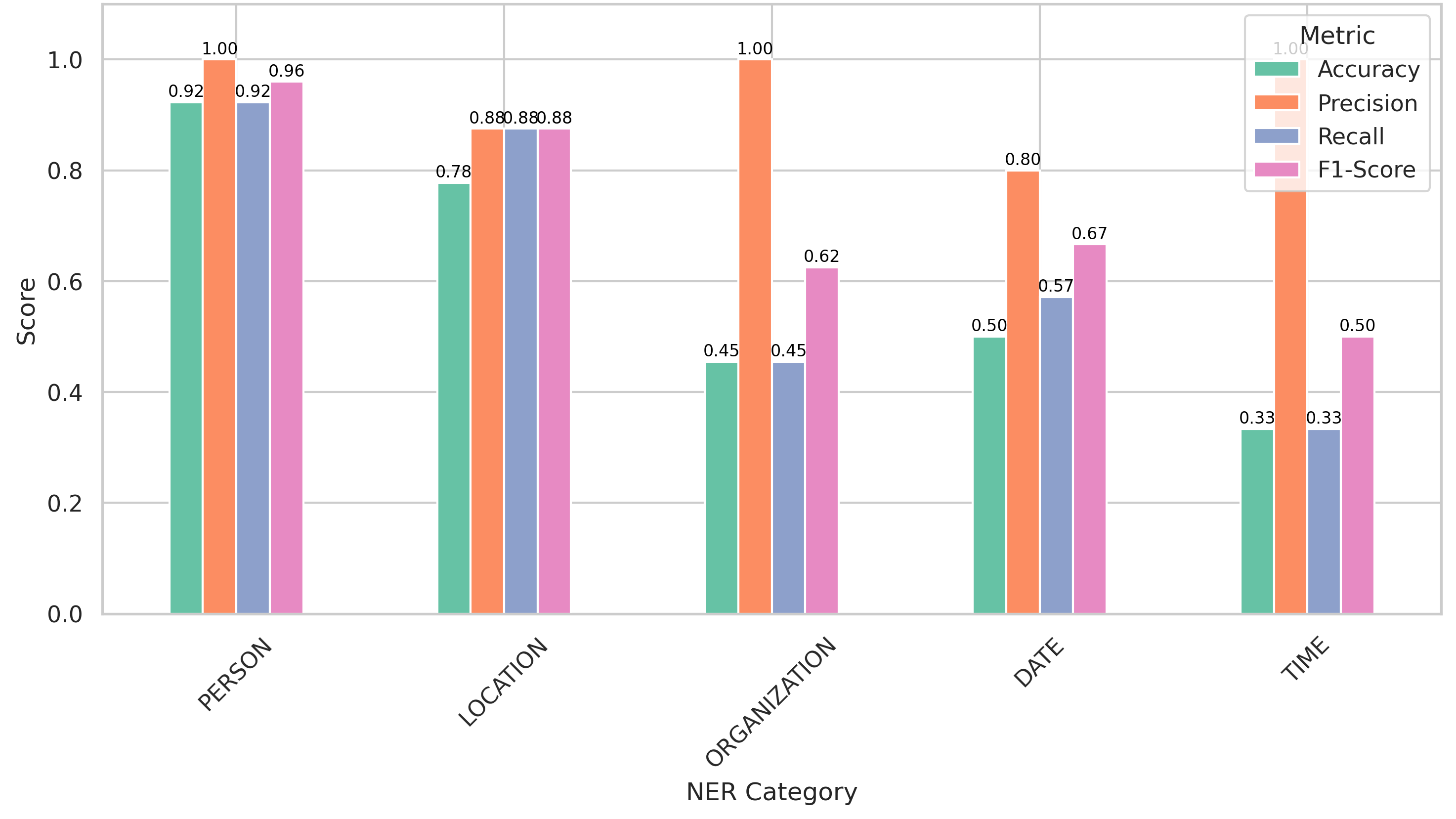} % Replace with your actual filename
  \caption{Per-Entity F1-Score for DeepSeek-V3 Across Five Entity Types}
  \label{fig:deepseek}
\end{figure}

DeepSeek closely mirrors Gemini’s performance in several areas. It matches Gemini’s near-perfect precision on PERSON and achieves a similar F1-score (0.960). On LOCATION, it also reaches an F1-score of 0.875, showing robust tagging of place names like Washington and Lincoln. It falls behind in ORGANIZATION tagging (F1 = 0.625), where it occasionally mislabels multi-word organizations or omits them entirely. Its performance on DATE (F1 = 0.667) and TIME (F1 = 0.500) is decent but less consistent than traditional models like spaCy or Stanza. These results indicate that while DeepSeek performs well on structured tags, it may benefit from more explicit prompt engineering when dealing with complex or nested entities (Figure~\ref{fig:deepseek}).

\begin{figure}
  \centering
      \includegraphics[width=10cm]{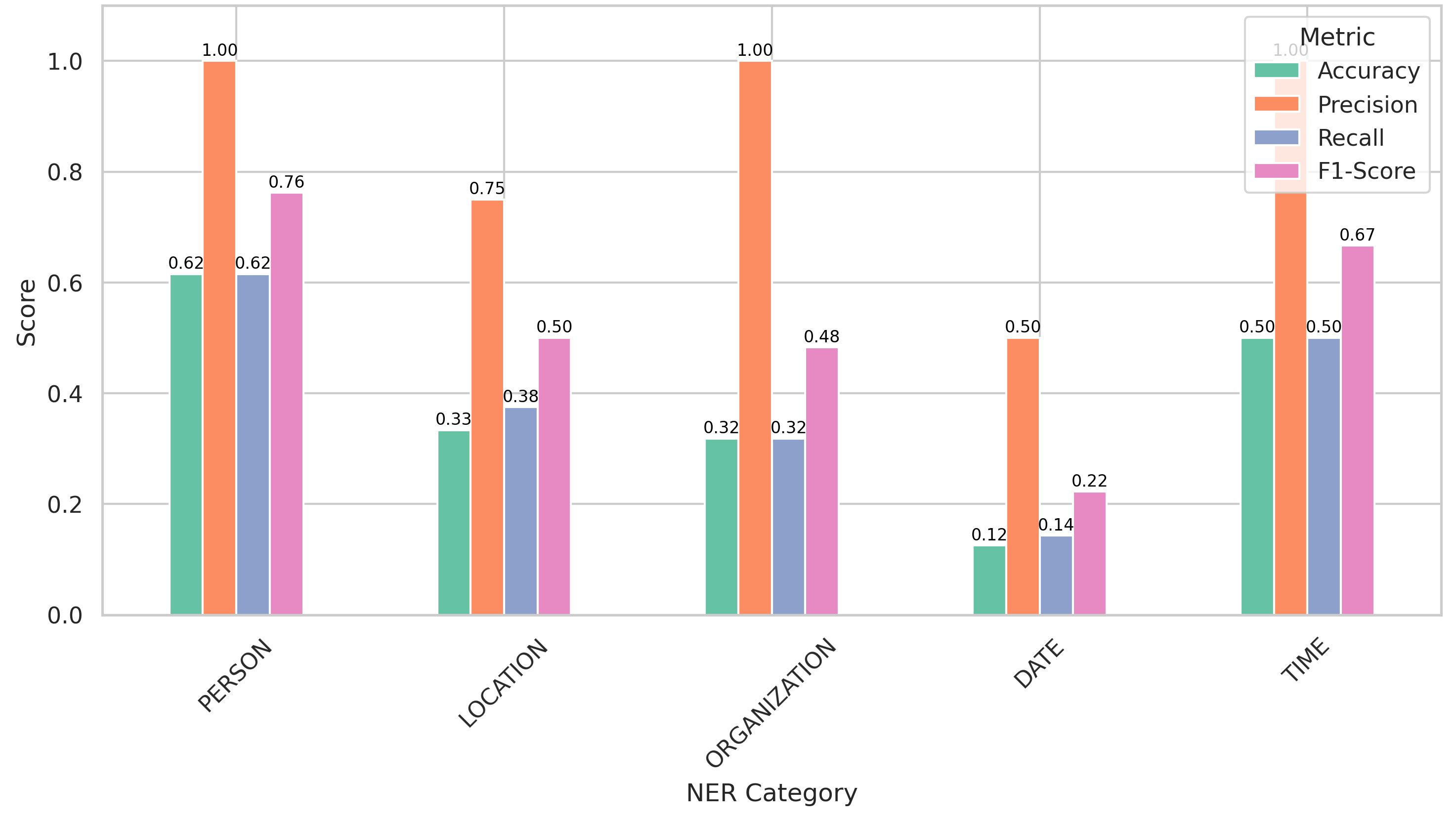} % Replace with your actual filename
  \caption{Per-Entity F1-Score for Qwen-3-4B Across Five Entity Types}
  \label{fig:qwen}
\end{figure}

Qwen shows mixed performance across entity types. It performs reasonably well on PERSON (F1 = 0.762), though it occasionally overpredicts this category. It struggles significantly with DATE (F1 = 0.222), often failing to recognize full date expressions like the fourth Friday of March. Its LOCATION performance (F1 = 0.500) suggests partial success, but it sometimes misclassifies spatial terms or misses them entirely. Qwen does perform relatively well on TIME (F1 = 0.667), correctly identifying expressions like dusk and midday. Overall, Qwen appears to be sensitive to context but inconsistent in its application of entity boundaries, leading to varied performance depending on the type of entity. (Figure~\ref{fig:qwen}).

\subsection{System-Level Performance Ranking (Macro-Averaged F1)}

\begin{figure}
  \centering
    \includegraphics[width=10cm]{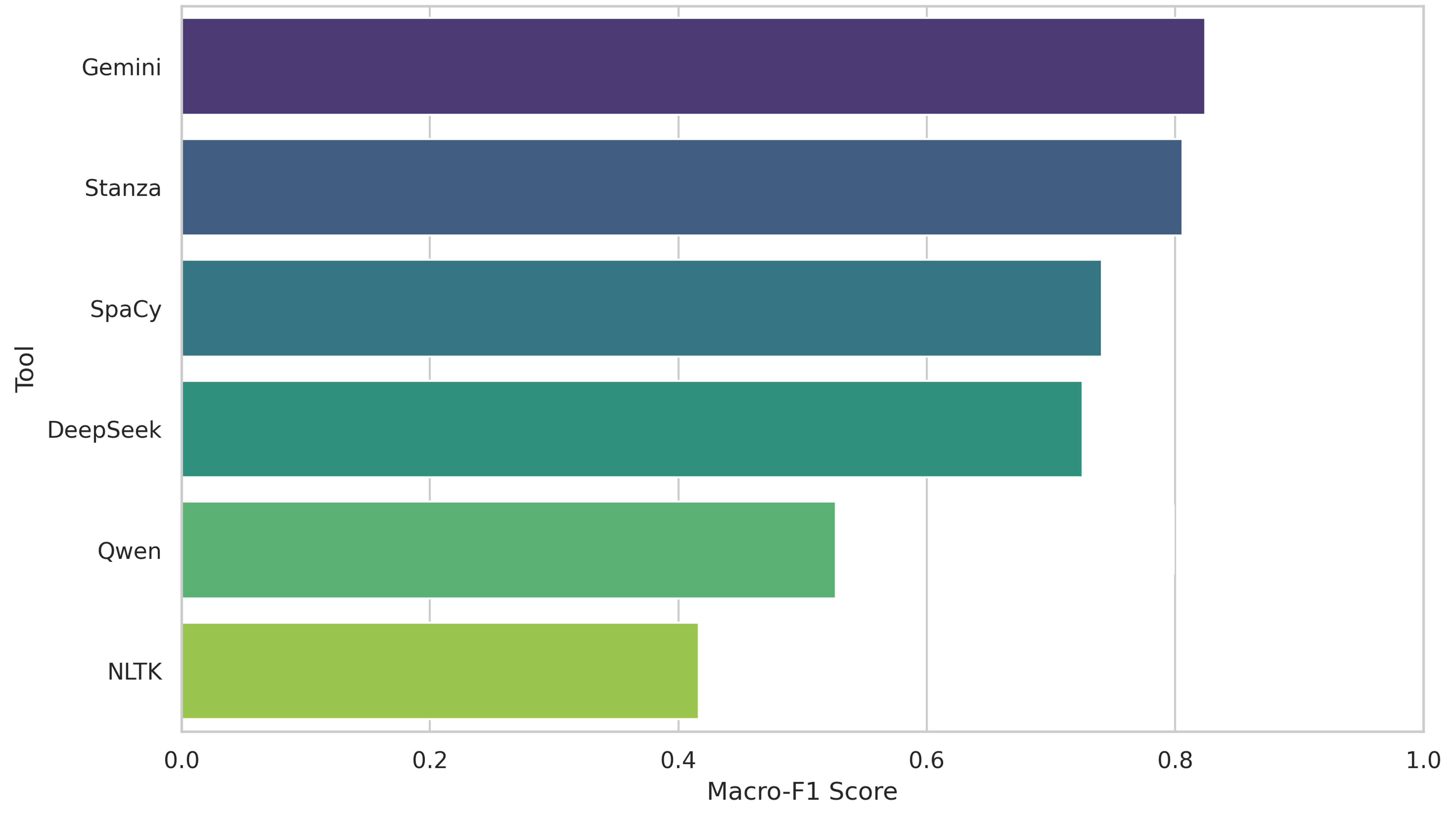} % Replace with your actual filename
  \caption{Macro-Averaged F1-Score by System (Higher = Better)}
  \label{fig:macro_avg}
\end{figure}

To provide an overall performance measure of each tool in Named Entity Recognition (NER), we computed the macro-averaged F1-score, which gives equal weight to each entity type. The results, as depicted in Figure~\ref{fig:macro_avg}, show that Gemini achieves the highest average F1-score of 0.824, indicating strong performance across most categories. Stanza follows closely with an average F1-score of 0.806, demonstrating robust NER capabilities. DeepSeek also performs well, achieving an average F1-score of 0.725.

In contrast, spaCy and DeepSeek show moderate performances with an average F1-score of 0.741 and 0.725 respectively. While Qwen achieves an average F1-score of 0.527, highlighting its competitive performance despite some variability, NLTK performs the weakest among traditional tools, with an average F1-score of 0.416, largely due to challenges in recognizing temporal expressions like DATE and TIME.

On this dataset, the results suggest that large language models (LLMs) generally outperform traditional NLP libraries, particularly in handling context-dependent entities such as PERSON. However, conventional tools like Stanza maintain strong consistency in structured tags such as LOCATION and DATE, making them suitable for applications where reliability is crucial.

\subsection{Group-Level Performance by Entity Type (LLMs vs. Traditional Tools)}

\begin{figure}
  \centering
  \includegraphics[width=10cm]{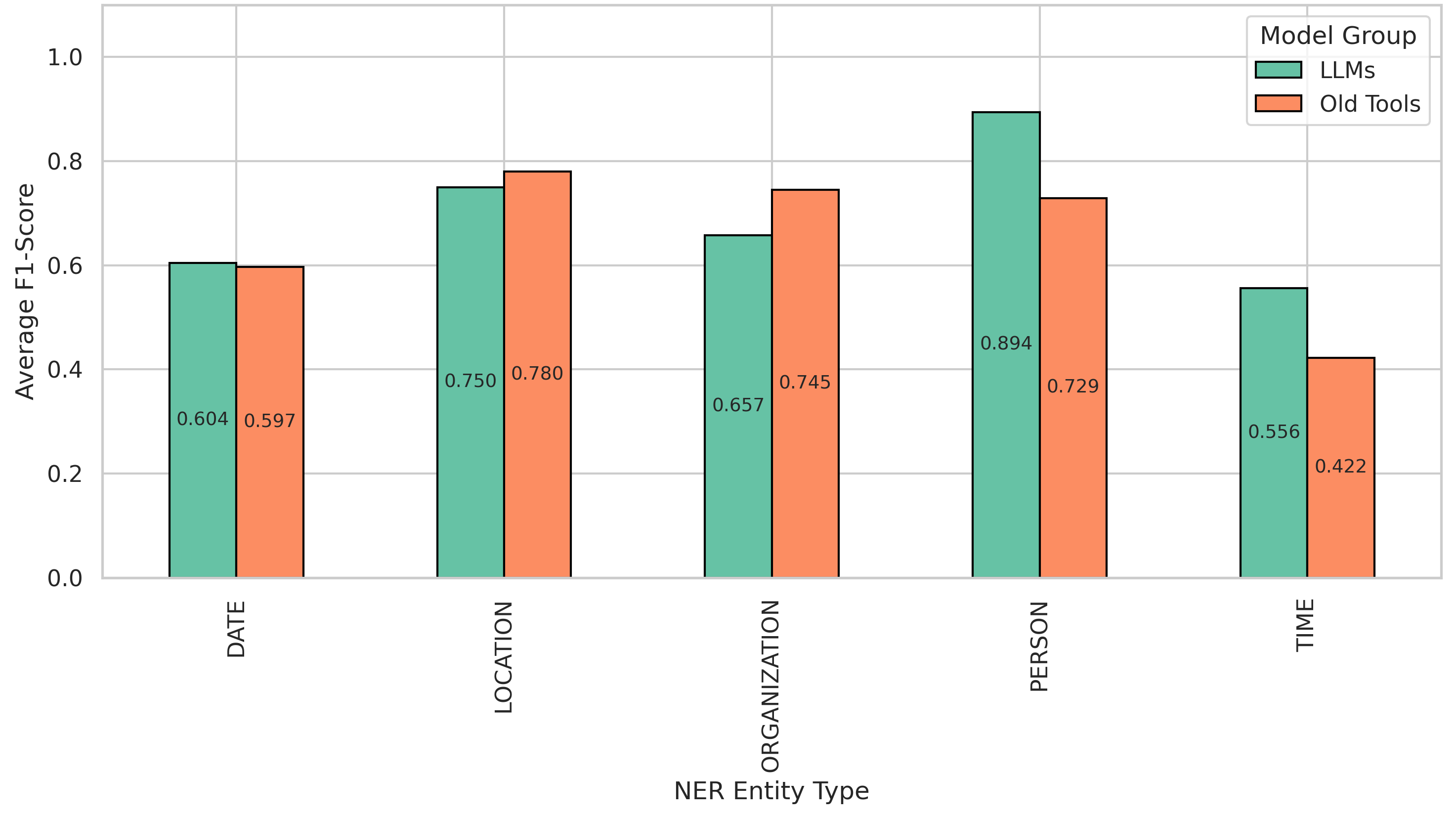} % Replace with your actual filename
  \caption{Average F1-Score by Entity Type: LLMs vs. Traditional Tools}
  \label{fig:group_comp}
\end{figure}

The average F1-scores across entity types (Figure~\ref{fig:group_comp}) reveal distinct performance patterns between traditional NLP tools and large language models. In the DATE category, LLMs achieved an average score of 0.604, slightly outperforming conventional Tools, which scored 0.597, indicating that both groups faced challenges in identifying temporal expressions like October or the fourth Friday of March. For LOCATION, both groups performed comparably, with LLMs scoring 0.750 and non-LLM Tools achieving 0.780, showing that traditional systems remain competitive in recognizing geographic names such as Washington or Madison Square. The most significant difference emerged in the PERSON category, where LLMs outperformed traditional Tools with an average F1-score of 0.923 compared to 0.692, highlighting their superior ability to recognize ambiguous person names like Justice and Hope. However, LLMs showed limitations in the ORGANIZATION category, scoring 0.657 versus 0.733 for traditional tools, suggesting that conventional Tools are more reliable when it comes to labeling institutional or company names. Both groups struggled similarly with TIME expressions, each scoring an average of 0.500, indicating room for improvement in handling time-of-day references like dusk or midday. These results suggest that while LLMs generally perform better in context-sensitive tagging tasks, especially for person names, traditional tools maintain consistency in structured tags such as LOCATION and ORGANIZATION.

\subsection{Aggregate Group Performance (LLMs vs. Traditional Tools)}

\begin{figure}
  \centering
  \includegraphics[width=10cm]{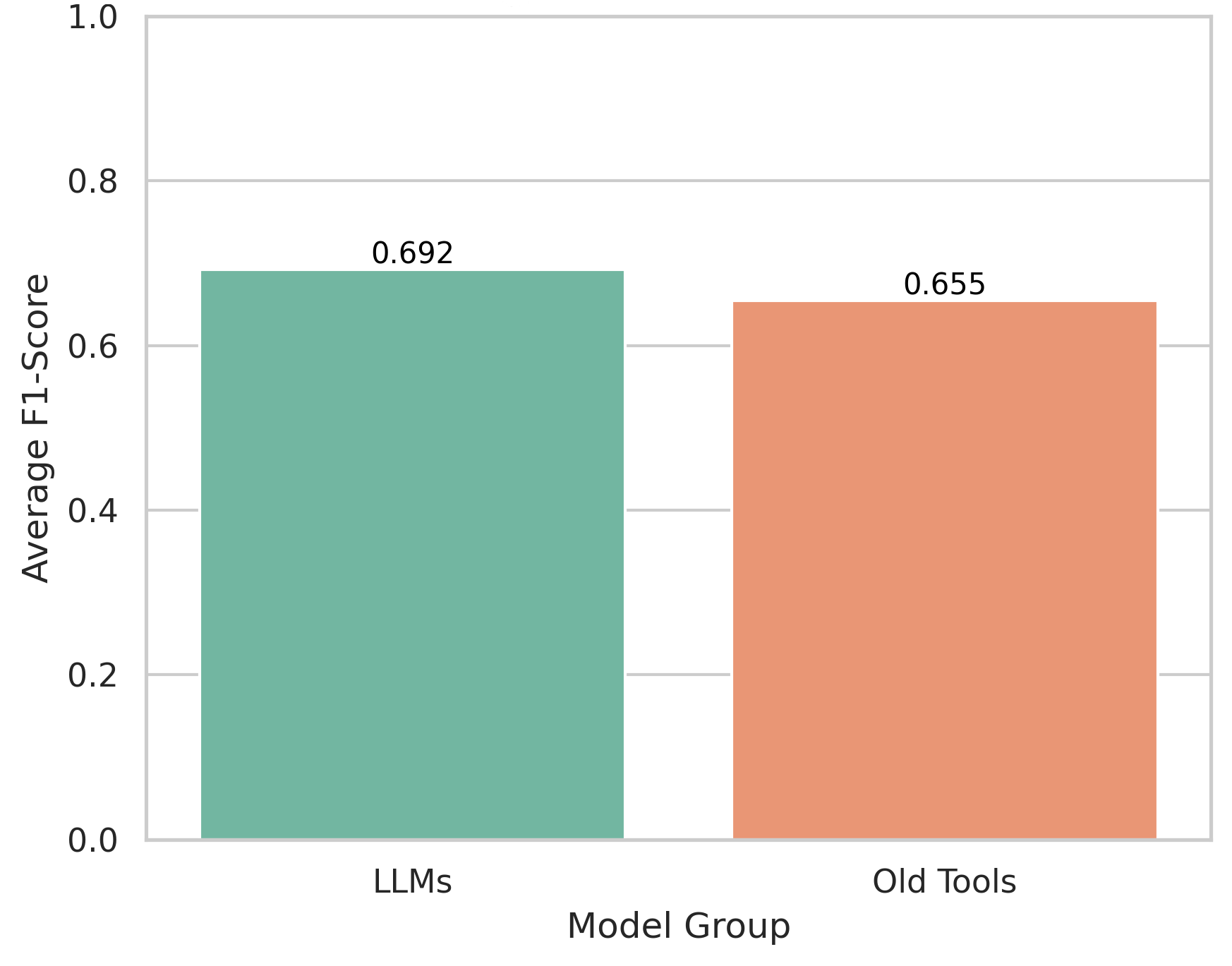} % Replace with your actual filename
  \caption{Aggregate F1-Score Comparison: LLMs vs. Traditional Tools}
  \label{fig:agg_comp}
\end{figure}

The overall performance of the models was evaluated by computing the average F1-score across all entity types. As shown in Figure~\ref{fig:agg_comp}, LLMs achieved an average F1-score of 0.692, while traditional Tools scored slightly lower with an average F1-score of 0.655. In this pilot study, LLMs demonstrated slightly higher average performance than traditional tools on our dataset. However, the difference is relatively small, suggesting that both groups perform comparably well overall. These results highlight the strengths of LLMs in handling complex or context-dependent entities, though traditional tools maintain robustness in structured tagging scenarios.

\section{Conclusion}
This pilot study evaluated six NER systems—three classical libraries (NLTK, spaCy, Stanza) and three large language models (Gemini, DeepSeek, Qwen)—on a 119-token set engineered for ambiguity. Gemini achieved the highest macro-F1 (0.824), driven by a 0.960 score on PERSON; it correctly interpreted “Justice Hope” as a person, confirming that contextual pre-training aids disambiguation of capitalized role words. DeepSeek matched Gemini on PERSON and LOCATION, yet its recall on multi-token organizations fell, showing that LLM gains are entity-specific. Qwen displayed wider fluctuation, underlining inter-model instability even at low temperature.

Among traditional tools, Stanza was the most robust (macro-F1 0.806). Its deep-learning pipeline, augmented by gazetteer features, delivered consistent results on LOCATION (0.857) and ORGANIZATION (0.846), and outperformed most systems on DATE (0.857). SpaCy excelled on DATE (0.933) but sacrificed recall on ambiguous persons, illustrating the precision–flexibility trade-off inherent in rule-supported taggers. NLTK lagged across all categories, failing entirely on temporal expressions, and thus serves best as a baseline rather than a production option.

Overall, LLMs surpass classical systems when context governs the decision, whereas libraries like Stanza remain preferable for high-volume, dictionary-driven spans where determinism and speed matter. In production pipelines that must process millions of documents under tight CPU or latency budgets, the lighter libraries still deliver the cheaper, reproducible solution; Gemini’s higher per-query cost is justified only when the input is rich in ambiguous names and recall outweighs runtime expense.

\subsection{Strengths and Limitations}
This study provides a focused investigation into named entity recognition (NER) ambiguity by comparing the performance of traditional NLP libraries (NLTK, spaCy, Stanza) and large language models (Gemini, DeepSeek, Qwen). A major strength lies in the creation of a manually annotated gold standard comprising 119 tokens across five entity types (PERSON, ORGANIZATION, LOCATION, DATE, and TIME). This fine-grained dataset enables a detailed comparison of token-level behavior across diverse modeling paradigms. Furthermore, by evaluating both traditional models and modern LLMs side by side, this work highlights complementary strengths and weaknesses, offering insights that extend beyond existing NER benchmarks.

Nonetheless, several limitations should be acknowledged. First, the dataset is small (119 tokens) and genre-specific, limiting statistical power, generalizability, and span-level evaluation scope. Second, annotation was performed by a single annotator due to time and resource constraints, also precluding calculation of inter-annotator agreement (IAA); consequently, no reliability score (e.g., Cohen’s kappa) can be reported. Future work should involve multiple annotators to ensure reliability and report agreement metrics such as Cohen’s kappa. Third, the use of API-based LLMs introduces two challenges: cost and output variability. Since LLM calls can be expensive, repeated runs for sensitivity analysis were limited; additionally, outputs may vary between calls, affecting reproducibility. Fourth, due to the proprietary nature of LLM APIs, exact reproducibility may be impacted by model updates over time. Moreover, temperatures were set to the lowest values offered by each API (Gemini $t = 0.2$, DeepSeek/Qwen $t = 0.1$); however, any temperature $> 0$ retains sampling randomness, so observed performance differences could partly reflect sampling variation rather than intrinsic model capability. Finally, this study relies on token-level evaluation, which, while appropriate for a pilot study, may penalize minor boundary errors (e.g., one missing token inside a multi-word organization), slightly underestimating true span-level F1. Future work should incorporate span-based evaluation metrics following established NER benchmarks.

Addressing these limitations opens clear avenues for future research: expanding the dataset size and domain diversity, involving multiple annotators to ensure reliability, conducting sensitivity analyses on LLM prompts and outputs, and releasing all annotated data and evaluation code to improve reproducibility.

\subsection{Future Work}
Future work should address the study’s primary limitations along multiple dimensions. First, the dataset should be extended both in size and domain diversity (news, social media, legal text) to evaluate whether the patterns observed here generalize beyond a concentrated ambiguity-rich passage. Second, annotation should be performed by multiple independent annotators to compute inter-annotator agreement (e.g., Cohen’s kappa) and to refine annotation guidelines; adjudication of disagreements would yield a more robust gold standard. Third, evaluations should report both token-level and span-level metrics, and include illustrative span examples to distinguish boundary errors from misclassification errors. Fourth, sensitivity analyses are required for LLMs: perform multiple runs per model, conduct temperature and prompt-engineering ablations (single-shot vs few-shot), and compare zero-shot prompts with light fine-tuning to quantify stability and cost-benefit tradeoffs. Fifth, release of the annotated dataset, the exact prompts, and the evaluation scripts (for example via a DOIed repository) will improve reproducibility and accelerate follow-up work; we recommend including a reproducible Docker or Colab notebook that reproduces the token alignment and scoring pipeline. Finally, future studies should evaluate runtime, memory, and monetary costs in addition to accuracy to provide practical guidelines for selecting models in production contexts and should explore cross-lingual and fairness analyses to assess whether similar ambiguity patterns arise across languages and demographic groups.

\subsection{Final Remarks}
While LLMs show promise in handling ambiguous or context-dependent entities, traditional tools remain competitive in structured tagging tasks. The choice between them should also consider speed, memory, API cost and reproducibility needs, not only small-scale scores.

\section*{Appendices}

\section*{A.1. Prompt and API Configuration: Gemini-1.5-flash}
Model checkpoint: gemini-1.5-flash-latest \\
Provider: Google Generative AI (google-generativeai Python SDK) \\
Date queried: 14 May 2025 \\
Temperature: 0.2 \\
Max tokens: not set (model decides) \\
Repetitions: single run

\vspace{1em}

\begin{lstlisting}
You are an expert in Named Entity Recognition (NER). Your task is to extract entities from the given text and tag each token as one of the following:
PERSON, LOCATION, ORGANIZATION, DATE, TIME, or O (if the token is not an entity).
Return the result as a list of rows. Each row must contain only the token and its NER tag, separated by a comma.
Format each row strictly as: Token,NER_Tag
Example:
Apple,ORGANIZATION
January,DATE
went,O
to,O
Paris,LOCATION
Do not include any introductory text, concluding remarks, or explanations. Only output the list of token-tag pairs.
\end{lstlisting}

\section*{A.2. Prompt and API Configuration: DeepSeek-V3}
Model checkpoint: deepseek/deepseek-chat:free \\
Provider: OpenRouter (\url{https://openrouter.ai/api/v1/chat/completions}) \\
Date queried: 14 May 2025 \\
Temperature: 0.1 \\
Max tokens: 1000 \\
Repetitions: single run

\textbf{System message:}
\vspace{1em}
\begin{lstlisting}
You are an expert Named Entity Recognition system. Your task is to identify entities in the provided text and return them in the specified JSON format.
\end{lstlisting}

\textbf{User message:}
\vspace{1em}
\begin{lstlisting}
Extract all named entities from the following text.
Identify entities such as PERSON, LOCATION, ORGANIZATION, DATE, TIME.
Return the result ONLY as a JSON list of objects. Each object in the list should have two keys:
1. "text": The exact extracted entity text.
2. "label": The entity type (e.g., "PERSON", "LOCATION", "ORGANIZATION").
For example, if the text is "Apple Inc. was founded by Steve Jobs in Cupertino on April 1, 1976.", the output should be:
[
  {"text": "Apple Inc.", "label": "ORGANIZATION"},
  {"text": "Steve Jobs", "label": "PERSON"},
  {"text": "Cupertino", "label": "LOCATION"},
  {"text": "April 1, 1976", "label": "DATE"}
]
If no entities are found, return an empty JSON list: [].
Do not include any explanations, introductory text, or any characters outside the JSON list itself.
\end{lstlisting}

\section*{A.3. Prompt and API Configuration: Qwen-3-4B}
Model checkpoint: qwen/qwen3-4b:free \\
Provider: OpenRouter (same endpoint as above) \\
Date queried: 14 May 2025 \\
Temperature: 0.1 \\
Max tokens: 1000 \\
Repetitions: single run

\textbf{System message:}
\vspace{1em}
\begin{lstlisting}
You are an expert Named Entity Recognition system. Your task is to identify entities in the provided text and return them in the specified JSON format.
\end{lstlisting}

\textbf{User message:}
\vspace{1em}
\begin{lstlisting}
Extract all named entities from the following text.
Identify entities such as PERSON, LOCATION, ORGANIZATION, DATE, TIME.
Return the result ONLY as a JSON list of objects. Each object in the list should have two keys:
1. "text": The exact extracted entity text.
2. "label": The entity type (e.g., "PERSON", "LOCATION", "ORGANIZATION").
For example, if the text is "Apple Inc. was founded by Steve Jobs in Cupertino on April 1, 1976.", the output should be:
[
  {"text": "Apple Inc.", "label": "ORGANIZATION"},
  {"text": "Steve Jobs", "label": "PERSON"},
  {"text": "Cupertino", "label": "LOCATION"},
  {"text": "April 1, 1976", "label": "DATE"}
]
If no entities are found, return an empty JSON list: [].
Do not include any explanations, introductory text, or any characters outside the JSON list itself.
\end{lstlisting}

\section*{B. Full Token-Level Evaluation Metrics by Tool and Entity Type}

\begin{table}[H]
\caption{Full Token-Level Evaluation Metrics by Tool and Entity Type}
\centering
\small
\setlength{\tabcolsep}{2pt}
\begin{tabular}{l l c c c c}
\toprule
\textbf{Tool} & \textbf{Entity} & \textbf{Accuracy} & \textbf{Precision} & \textbf{Recall} & \textbf{F1-Score} \\
\midrule
NLTK & PERSON & 0.733333 & 0.846154 & 0.846154 & 0.846154 \\
NLTK & LOCATION & 0.454545 & 0.625 & 0.625 & 0.625 \\
NLTK & ORGANIZATION & 0.44 & 0.785714 & 0.5 & 0.611111 \\
NLTK & DATE & 0 & 0 & 0 & 0 \\
NLTK & TIME & 0 & 0 & 0 & 0 \\
\midrule
spaCy & PERSON & 0.307692 & 1 & 0.307692 & 0.470588 \\
spaCy & LOCATION & 0.75 & 1 & 0.75 & 0.857143 \\
spaCy & ORGANIZATION & 0.636364 & 0.65625 & 0.954545 & 0.777778 \\
spaCy & DATE & 0.875 & 0.875 & 1 & 0.933333 \\
spaCy & TIME & 0.5 & 1 & 0.5 & 0.666667 \\
\midrule
Stanza & PERSON & 0.769231 & 1 & 0.769231 & 0.869565 \\
Stanza & LOCATION & 0.75 & 1 & 0.75 & 0.857143 \\
Stanza & ORGANIZATION & 0.733333 & 0.733333 & 1 & 0.846154 \\
Stanza & DATE & 0.75 & 0.857143 & 0.857143 & 0.857143 \\
Stanza & TIME & 0.428571 & 0.75 & 0.5 & 0.6 \\
\midrule
Gemini & PERSON & 0.923077 & 1 & 0.923077 & 0.96 \\
Gemini & LOCATION & 0.777778 & 0.875 & 0.875 & 0.875 \\
Gemini & ORGANIZATION & 0.76 & 0.863636 & 0.863636 & 0.863636 \\
Gemini & DATE & 0.857143 & 1 & 0.857143 & 0.923077 \\
Gemini & TIME & 0.333333 & 1 & 0.333333 & 0.5 \\
\midrule
DeepSeek & PERSON & 0.923077 & 1 & 0.923077 & 0.96 \\
DeepSeek & LOCATION & 0.777778 & 0.875 & 0.875 & 0.875 \\
DeepSeek & ORGANIZATION & 0.454545 & 1 & 0.454545 & 0.625 \\
DeepSeek & DATE & 0.5 & 0.8 & 0.571429 & 0.666667 \\
DeepSeek & TIME & 0.333333 & 1 & 0.333333 & 0.5 \\
\midrule
Qwen & PERSON & 0.615385 & 1 & 0.615385 & 0.761905 \\
Qwen & LOCATION & 0.333333 & 0.75 & 0.375 & 0.5 \\
Qwen & ORGANIZATION & 0.318182 & 1 & 0.318182 & 0.482759 \\
Qwen & DATE & 0.125 & 0.5 & 0.142857 & 0.222222 \\
Qwen & TIME & 0.5 & 1 & 0.5 & 0.666667 \\
\bottomrule
\end{tabular}
\label{tab:full_metrics}
\end{table}

\section*{C. Annotated Dataset: Input Text, Tokenization, and Gold Labels}

\scriptsize
\setlength{\tabcolsep}{2pt}
\begin{longtable}{r l l l l l l l l}
\caption{Annotated Dataset: Token-Level Alignment and Human Annotations} \\
\toprule
\textbf{ID} & \textbf{Token} & \textbf{NLTK} & \textbf{spaCy} & \textbf{Stanza} & \textbf{Gemini} & \textbf{DS-V3} & \textbf{Qwen} & \textbf{Gold} \\
\midrule
\endfirsthead

\multicolumn{9}{c}{{\tablename\ \thetable{} - continued from previous page}} \\
\toprule
\textbf{ID} & \textbf{Token} & \textbf{NLTK} & \textbf{spaCy} & \textbf{Stanza} & \textbf{Gemini} & \textbf{DS-V3} & \textbf{Qwen} & \textbf{Gold} \\
\midrule
\endhead

\midrule
\multicolumn{9}{r}{{Continued on next page}} \\
\endfoot

\bottomrule
\endlastfoot

1 & On & O & O & O & O & N/A & N/A & O \\
2 & a & O & O & O & O & N/A & N/A & O \\
3 & chilly & O & O & O & O & N/A & N/A & O \\
4 & Thursday & O & DATE & TIME & DATE & N/A & N/A & DATE \\
5 & morning & O & TIME & TIME & O & N/A & N/A & TIME \\
6 & , & O & O & O & O & N/A & N/A & O \\
7 & April & PERSON & PERSON & PERSON & PERSON & PERSON & N/A & PERSON \\
8 & Blake & PERSON & PERSON & PERSON & PERSON & PERSON & N/A & PERSON \\
9 & met & O & O & O & O & N/A & N/A & O \\
10 & Justice & ORGANIZATION & O & PERSON & PERSON & PERSON & N/A & PERSON \\
11 & Hope & O & O & PERSON & PERSON & PERSON & N/A & PERSON \\
12 & at & O & O & O & O & N/A & N/A & O \\
13 & Madison & LOCATION & LOCATION & LOCATION & LOCATION & LOCATION & N/A & LOCATION \\
14 & Square & LOCATION & LOCATION & LOCATION & LOCATION & LOCATION & N/A & LOCATION \\
15 & , & O & O & O & O & N/A & N/A & O \\
16 & near & O & O & O & O & N/A & N/A & O \\
17 & Washington & LOCATION & LOCATION & LOCATION & LOCATION & LOCATION & N/A & LOCATION \\
18 & Tower & LOCATION & LOCATION & LOCATION & LOCATION & LOCATION & N/A & LOCATION \\
19 & . & O & O & O & O & N/A & N/A & O \\
20 & At & O & O & O & O & N/A & N/A & O \\
21 & 8:45 & O & TIME & TIME & TIME & TIME & TIME & TIME \\
22 & , & O & O & O & O & N/A & N/A & O \\
23 & a & O & O & O & O & N/A & N/A & O \\
24 & memo & O & O & O & O & N/A & N/A & O \\
25 & from & O & O & O & O & N/A & N/A & O \\
26 & the & O & ORGANIZATION & ORGANIZATION & ORGANIZATION & N/A & N/A & O \\
27 & Center & ORGANIZATION & ORGANIZATION & ORGANIZATION & ORGANIZATION & ORGANIZATION & ORGANIZATION & ORGANIZATION \\
28 & for & O & ORGANIZATION & ORGANIZATION & ORGANIZATION & N/A & N/A & ORGANIZATION \\
29 & Civic & PERSON & ORGANIZATION & ORGANIZATION & ORGANIZATION & N/A & N/A & ORGANIZATION \\
30 & Leadership & PERSON & ORGANIZATION & ORGANIZATION & ORGANIZATION & ORGANIZATION & N/A & ORGANIZATION \\
31 & reached & O & O & O & O & N/A & N/A & O \\
32 & the & O & O & O & O & N/A & N/A & O \\
33 & office & O & O & O & O & N/A & N/A & O \\
34 & of & O & O & O & O & N/A & N/A & O \\
35 & River & PERSON & ORGANIZATION & PERSON & PERSON & PERSON & PERSON & PERSON \\
36 & Clark & PERSON & ORGANIZATION & PERSON & PERSON & PERSON & PERSON & PERSON \\
37 & , & O & O & O & O & N/A & N/A & O \\
38 & head & O & O & O & O & N/A & N/A & O \\
39 & of & O & O & O & O & N/A & N/A & O \\
40 & Northern & LOCATION & ORGANIZATION & ORGANIZATION & ORGANIZATION & ORGANIZATION & ORGANIZATION & ORGANIZATION \\
41 & Union & LOCATION & ORGANIZATION & ORGANIZATION & ORGANIZATION & N/A & N/A & ORGANIZATION \\
42 & Trust & LOCATION & ORGANIZATION & ORGANIZATION & ORGANIZATION & N/A & N/A & ORGANIZATION \\
43 & . & O & O & O & O & N/A & N/A & O \\
44 & Later & O & O & O & O & N/A & N/A & O \\
45 & , & O & O & O & O & N/A & N/A & O \\
46 & Horizon & ORGANIZATION & ORGANIZATION & ORGANIZATION & ORGANIZATION & ORGANIZATION & ORGANIZATION & ORGANIZATION \\
47 & filed & O & O & O & O & N/A & N/A & O \\
48 & a & O & O & O & O & N/A & N/A & O \\
49 & complaint & O & O & O & O & N/A & N/A & O \\
50 & dated & O & O & O & O & N/A & N/A & O \\
51 & October & O & DATE & DATE & DATE & DATE & DATE & DATE \\
52 & 2018 & O & DATE & DATE & DATE & DATE & N/A & DATE \\
53 & with & O & O & O & O & N/A & N/A & O \\
54 & the & O & ORGANIZATION & ORGANIZATION & ORGANIZATION & N/A & N/A & O \\
55 & Department & ORGANIZATION & ORGANIZATION & ORGANIZATION & ORGANIZATION & ORGANIZATION & ORGANIZATION & ORGANIZATION \\
56 & of & O & ORGANIZATION & ORGANIZATION & O & N/A & N/A & ORGANIZATION \\
57 & Historical & ORGANIZATION & ORGANIZATION & ORGANIZATION & ORGANIZATION & N/A & N/A & ORGANIZATION \\
58 & Records & ORGANIZATION & ORGANIZATION & ORGANIZATION & ORGANIZATION & N/A & N/A & ORGANIZATION \\
59 & . & O & O & O & O & N/A & N/A & O \\
60 & By & O & O & O & O & N/A & N/A & O \\
61 & midday & O & TIME & TIME & TIME & TIME & TIME & TIME \\
62 & , & O & O & O & O & N/A & N/A & O \\
63 & journalists & O & O & O & O & N/A & N/A & O \\
64 & from & O & O & O & O & N/A & N/A & O \\
65 & The & ORGANIZATION & ORGANIZATION & ORGANIZATION & ORGANIZATION & N/A & N/A & ORGANIZATION \\
66 & Chronicle & ORGANIZATION & ORGANIZATION & ORGANIZATION & ORGANIZATION & ORGANIZATION & ORGANIZATION & ORGANIZATION \\
67 & and & O & O & ORGANIZATION & O & N/A & N/A & ORGANIZATION \\
68 & Liberty & ORGANIZATION & ORGANIZATION & ORGANIZATION & ORGANIZATION & N/A & N/A & ORGANIZATION \\
69 & Press & O & ORGANIZATION & ORGANIZATION & ORGANIZATION & N/A & N/A & ORGANIZATION \\
70 & were & O & O & O & O & N/A & N/A & O \\
71 & circling & O & O & O & O & N/A & N/A & O \\
72 & City & ORGANIZATION & LOCATION & LOCATION & LOCATION & LOCATION & LOCATION & LOCATION \\
73 & Hall & ORGANIZATION & LOCATION & LOCATION & LOCATION & LOCATION & LOCATION & LOCATION \\
74 & . & O & O & O & O & N/A & N/A & O \\
75 & Jordan & PERSON & PERSON & PERSON & PERSON & PERSON & PERSON & PERSON \\
76 & Reed & PERSON & PERSON & PERSON & PERSON & PERSON & PERSON & PERSON \\
77 & mentioned & O & O & O & O & N/A & N/A & O \\
78 & a & O & O & O & O & N/A & N/A & O \\
79 & meeting & O & O & O & O & N/A & N/A & O \\
80 & on & O & O & O & O & N/A & N/A & O \\
81 & the & O & DATE & DATE & O & DATE & DATE & O \\
82 & fourth & O & DATE & DATE & DATE & DATE & N/A & DATE \\
83 & Friday & O & DATE & DATE & DATE & DATE & N/A & DATE \\
84 & of & O & DATE & DATE & O & N/A & N/A & DATE \\
85 & March & O & DATE & DATE & DATE & N/A & N/A & DATE \\
86 & . & O & O & O & O & N/A & N/A & O \\
87 & Meanwhile & O & O & O & O & N/A & N/A & O \\
88 & , & O & O & O & O & N/A & N/A & O \\
89 & Trinity & PERSON & ORGANIZATION & PERSON & PERSON & PERSON & PERSON & PERSON \\
90 & Wells & PERSON & ORGANIZATION & PERSON & PERSON & PERSON & PERSON & PERSON \\
91 & , & O & O & O & O & N/A & N/A & O \\
92 & formerly & O & O & O & O & N/A & N/A & O \\
93 & of & O & O & O & O & N/A & N/A & O \\
94 & Bridgewater & ORGANIZATION & ORGANIZATION & ORGANIZATION & ORGANIZATION & ORGANIZATION & ORGANIZATION & ORGANIZATION \\
95 & School & ORGANIZATION & ORGANIZATION & ORGANIZATION & ORGANIZATION & ORGANIZATION & N/A & ORGANIZATION \\
96 & , & O & O & O & O & N/A & N/A & O \\
97 & was & O & O & O & O & N/A & N/A & O \\
98 & seen & O & O & O & O & N/A & N/A & O \\
99 & near & O & O & O & O & N/A & N/A & LOCATION \\
100 & Lincoln & LOCATION & ORGANIZATION & ORGANIZATION & LOCATION & LOCATION & LOCATION & LOCATION \\
101 & . & O & O & O & O & N/A & N/A & O \\
102 & Just & O & O & O & O & N/A & N/A & TIME \\
103 & before & O & O & O & O & N/A & N/A & TIME \\
104 & dusk & O & O & O & O & N/A & TIME & TIME \\
105 & , & O & O & O & O & N/A & N/A & O \\
106 & the & O & ORGANIZATION & ORGANIZATION & ORGANIZATION & N/A & N/A & O \\
107 & Mayor & GPE & ORGANIZATION & ORGANIZATION & ORGANIZATION & ORGANIZATION & ORGANIZATION & ORGANIZATION \\
108 & ’s & O & ORGANIZATION & ORGANIZATION & O & N/A & N/A & ORGANIZATION \\
109 & Office & ORGANIZATION & ORGANIZATION & ORGANIZATION & ORGANIZATION & ORGANIZATION & N/A & ORGANIZATION \\
110 & received & O & O & O & O & N/A & N/A & O \\
111 & a & O & O & O & O & N/A & N/A & O \\
112 & tip & O & O & O & O & N/A & N/A & O \\
113 & linked & O & O & O & O & N/A & N/A & O \\
114 & to & O & O & O & O & N/A & N/A & O \\
115 & Ashley & PERSON & ORGANIZATION & ORGANIZATION & PERSON & PERSON & PERSON & PERSON \\
116 & Fields & PERSON & ORGANIZATION & ORGANIZATION & PERSON & PERSON & PERSON & PERSON \\
117 & and & O & O & ORGANIZATION & O & N/A & N/A & O \\
118 & Pine & PERSON & ORGANIZATION & ORGANIZATION & LOCATION & LOCATION & LOCATION & PERSON \\
119 & . & O & O & O & O & N/A & N/A & O \\
\end{longtable}

\bibliographystyle{unsrt}

\end{document}